\documentclass{article}

\usepackage{times}
\usepackage{eqnarray}
\usepackage{graphicx} 
\usepackage{subfigure} 
\usepackage{natbib}
\usepackage{amsfonts}
\usepackage{array}

\usepackage{algorithm}
\usepackage{algorithmic}
\usepackage{tabularx}
\usepackage{tabulary}

\usepackage{hyperref}

\usepackage[nohyperref, accepted]{icml2017} 

\icmltitlerunning{Unsupervised Domain Adaptation Using Approximate Label Matching}

\begin{document} 

\twocolumn[
\icmltitle{Unsupervised Domain Adaptation Using Approximate Label Matching}


\begin{icmlauthorlist}
\icmlauthor{Jordan T. Ash}{princeton}
\icmlauthor{Robert E. Schapire}{msr}
\icmlauthor{Barbara E. Engelhardt}{princeton}
\end{icmlauthorlist}

\icmlaffiliation{princeton}{Princeton University, Princeton, NJ, USA}
\icmlaffiliation{msr}{Microsoft Research, New York, NY, USA}

\icmlcorrespondingauthor{Jordan T. Ash}{jordanta@cs.princeton.edu}

\icmlkeywords{domain adaptation, neural networks, deep learning, generative adversarial networks}

\vskip 0.3in
]
\printAffiliationsAndNotice{}

\begin{abstract} 
Domain adaptation addresses the problem created when training data is generated by a so-called source distribution, but test data is generated by a significantly different target distribution. In this work, we present approximate label matching (ALM), a new unsupervised domain adaptation technique that creates and leverages a rough labeling on the test samples, then uses these noisy labels to learn a transformation that aligns the source and target samples.
We show that the transformation estimated by ALM has favorable properties compared to transformations estimated by other methods, which do not use any kind of target labeling. Our model is regularized by requiring that a classifier trained to discriminate source from transformed target samples cannot distinguish between the two. We experiment with ALM on simulated and real data, and show that it outperforms techniques commonly used in the field.

\end{abstract}

\section{Introduction}
\label{section:intro}
Intuitively, intelligent agents should be able to improve on one task after having learned a similar kind of task. A human, for example, might be more capable of understanding Italian after having learned Spanish, or of playing tennis after having learned to play badminton. The goal of \emph{domain adaptation} is to endow machine intelligence with this same sort of capability.

In the usual classification or regression framework, we assume that training and test data are generated by the same underlying distribution. When this assumption does not hold, test performance can be significantly worse than training performance. This problem comes up in many areas of machine learning, particularly in natural language understanding \cite{nlpBengio} and computer vision \cite{vis}. As an example, we may want to build a general-purpose sentiment classifier, but only have access to highly-biased training data, such as book reviews from an online store. Similarly, a model trained on photos taken by a webcam will likely generalize poorly to photos taken by a higher-resolution DSLR camera.

The ability to improve the generalizability of fitted models is crucial to the real-world effectiveness of data-hungry methods like deep learning. Domain adaptation holds the promise of allowing these models to be fitted on datasets where labeled examples are abundant, then used to make predictions on a separate dataset where labeled samples are much more scarce, and that may be generated by a distribution different than that of the samples on which the network was originally trained. Domain adaptation has also been connected to a wide array of important machine learning tasks, including counterfactual inference \cite{sontag} and off-policy reinforcement learning \cite{rl}. 

Domain adaptation comes in three specific forms. In supervised domain adaptation, we are given several fully-labeled sources, and the goal is to learn a model that is stronger than one trained on any of the sources alone. In semi-supervised domain adaptation, we again have fully-labeled sources, but also a partially-labeled target domain. In this setting we are interested in classifying the unlabeled target data well by making use of all available information. In this article, we consider unsupervised domain adaptation, where we are given labeled source examples but do not have access to labels on target examples.

To coerce a model trained on one domain into performing well on another, domain adaptation methods often learn a transformation that makes the source samples statistically similar to the target samples. Correspondingly, the difficulty of domain adaptation lies in learning a high-quality transformation. Previous methods typically learn this transformation by considering only source and target samples, and sometimes source labels, but do not make any use of the target labels since they are not provided in this version of the problem.

In this work, we propose a new approach to unsupervised domain adaptation called approximate label matching (ALM). Our method constructs a rough labeling of the target data that we then exploit to dramatically improve the quality of the learned transformation between source and target data. 

\begin{figure}[h]
\vskip 0.2in
\begin{center}
\centerline{\includegraphics[trim={0cm 2.2cm 0cm 2.2cm}, clip, scale=0.2]{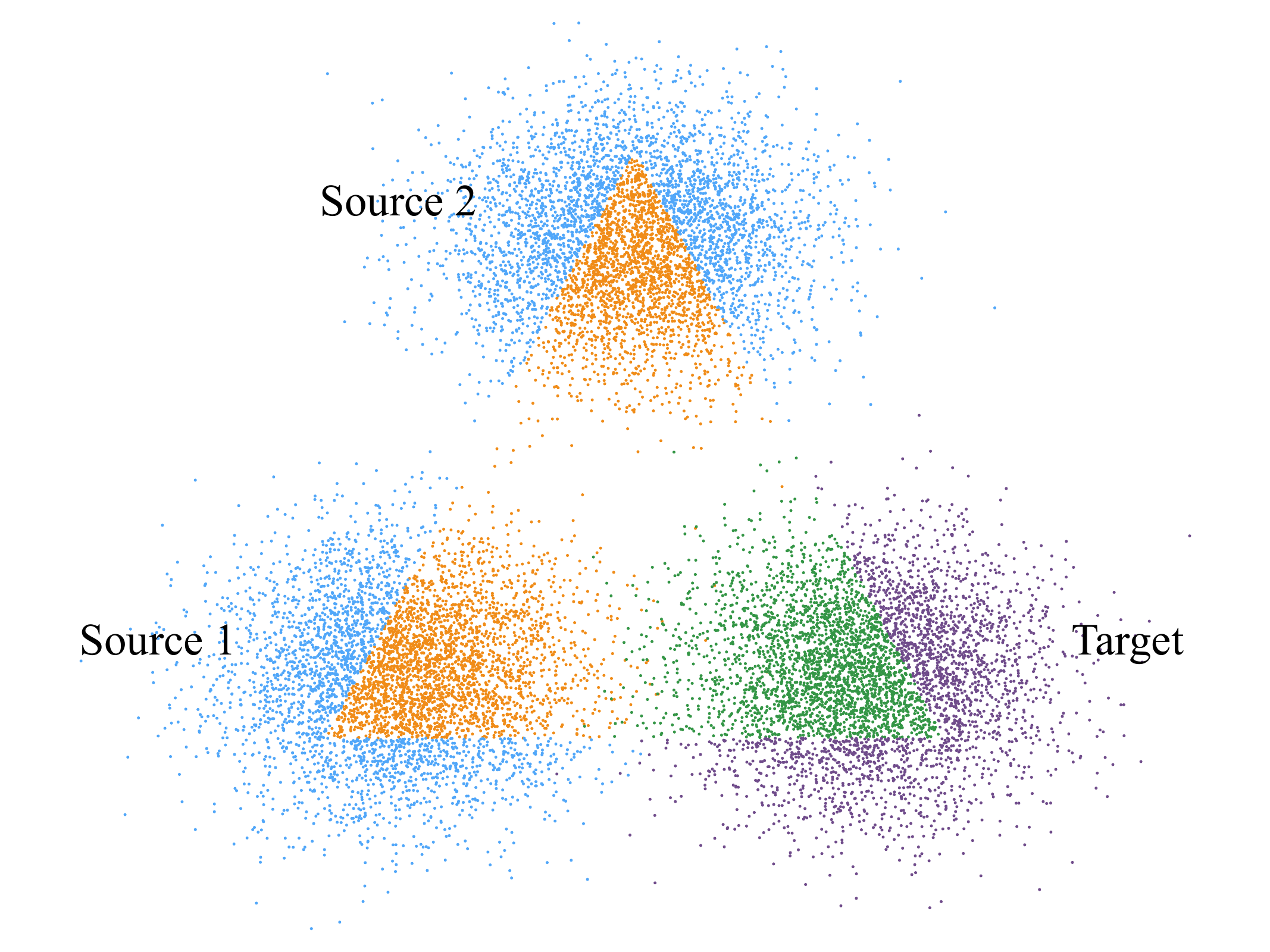}}
\caption{\textbf{The synthetic dataset, which includes two sources and one target.} Blue points are positive source samples, orange points are negative source samples, and purple and green points are positive and negative target samples respectively. Target domain labels are shown here for clarity, but the classifier does not have access to labels from the target dataset.}
\label{ex0}
\end{center}
\vskip -0.2in
\end{figure}

As a concrete illustration, we show a synthetic unsupervised domain adaptation problem, discussed further in Section~\ref{section:synth}, that highlights the potential importance of target label information in domain adaptation tasks (Figure~\ref{ex0}). We place three circular domains on each of the three vertices of an equilateral triangle. Points that are inside the triangle's perimeter are assigned negative labels and those outside are assigned positive labels. Despite the obvious similarities of each domain, without somehow incorporating target labels, it would clearly be impossible to uncover the rotation and translation that aligns the target data to either of the available sources. We find that ALM performs well in this adversarial situation, while other methods struggle. This example might seem contrived, but we show in experiments (Section~\ref{section:meg}) that real data can have similar properties.\looseness=-1

In Sections~\ref{our method} and \ref{sec:roughLabeling} we introduce and provide an intuitive justification for our method. In Section~\ref{sec:adreg} we discuss what it means for our model to overfit, and provide a simple regularization scheme that prevents this behavior. Finally, in Sections~\ref{section:synth} and \ref{section:results} we present comparative results of ALM applied to both simulated and real data, and we show that ALM outperforms related methods for domain adaptation.

\subsection{Related Work}

Domain adaptation has been approached in two distinct ways. In the first, training samples are re-weighted to make the resulting hypothesis better suited for classification on the test set. Kernel Mean Matching (KMM) is an example of a domain adaptation technique that falls into this category \cite{kmm}. KMM re-weights source points in an effort to make the means of the source and target datasets as close as possible in a reproducing kernel Hilbert space, which circumvents the need to approximate source or target distributions. Like other kernel methods, a vanilla implementation of KMM requires the construction of a matrix that is square with the number of samples in the target dataset, causing potential difficulty when working with large amounts of data.

In the alternative approach, one or both of the domains are transformed into a space where they better match each other. Once in this new representation, a classifier trained on a source is expected to perform well on the unlabeled target samples. A simple and effective algorithm that takes this approach is Subspace Alignment (SA), which finds a matrix that minimizes the Bregman divergence between the transformed source and target \cite{SA}. The PCA-based solution for SA is extremely efficient.

Unlike ALM, SA and KMM do not make use of label information from either the source or target datasets. Like SA, ALM learns a transformation to match the source and target domain, but SA is restricted to learning linear transformations, whereas ALM is able to estimate nonlinear transformations. \looseness=-1

More recently, several deep learning-based solutions to domain adaptation have been proposed \cite{mtae, dannJmlr}. One popular example of this is domain adversarial neural networks (DANNs), which consist of three parts: a transformation module, label classification module, and domain classification module \cite{dannJmlr}. Their goal, like with SA, is to uncover a transformation that projects both the source and target data into a canonical space where a model trained on the source data will generalize well to the unlabeled target. The transformation module embeds both the source and target datasets, and the label classification module uses that representation to distinguish between labeled source samples. The domain classification module is attached to the transformation module via a gradient-reversal layer, and is trained to distinguish between source and target examples. Gradient reversal causes the transformation module to learn to confuse the domain classification module, ideally forcing the transformed source and transformed target data to have similar distributions. All of the DANN modules are trained simultaneously. \looseness=-1

The ALM architecture consists of three neural network modules that are similar to those used in DANNs, but the two methods differ in a number of important ways. A significant difference between the two procedures is our use of a rough, better-than-guessing hypothesis on the target samples, which allows ALM to uncover transformations that other techniques cannot. A subtle difference between the two methods is that the transformation learned by ALM is applied only to the target, not the source, which we find is important for preventing overfitting in ALM. Also, unlike DANNs, our classification module is trained offline on the source dataset, and we make use of a true discriminator (discussed in Section~\ref{sec:adreg}) rather than a gradient reversal layer, because gradient reversal can cause instability while training \cite{martin}. 

Approximate label matching (ALM) is somewhat reminiscent of co-training, which was developed as a method for classifying data when there are few training examples available \cite{cotraining}. The idea is to create two different hypotheses by training two classifiers on two different representations of the same data. Test samples on which the two hypotheses strongly agree are added to the training set, and the process is repeated. Like co-training, ALM takes advantage of two candidate hypotheses: a rough labeling that is obtained offline, and an alternate labeling created by passing transformed target samples through a classifier trained on the source dataset.

\section{Approximate Label Matching (ALM)}
\label{our method}

In unsupervised domain adaptation, the predictor has been given one or more source datasets $(X^s, Y^s)$, $s = 1, \ldots, k$ and one target sequence $X^\star = \langle x^\star_1, \ldots, x^\star_n \rangle$. Each source dataset $s$ includes an input sequence $X^s = \langle x^s_1, \ldots, x^s_{m_s} \rangle$ and a label sequence $Y^s = \langle y^s_1, \ldots, y^s_{m_s} \rangle$, each with $m_s$ elements. In a binary classification problem, every $y^s_i \in \{0, 1\}$ corresponds to a label for a $d$-dimensional input $x^s_i \in \mathbb{R}^d$. For each $x^{\star}_{i} \in \mathbb{R}^d$ in the target sequence, the goal of the classifier is to predict the (unknown) label $y^\star_i \in \{0, 1\}$. Throughout this article, we use notation like $f_s(X^s)$ to denote the sequence of predictions produced by $f_s$ on all $x^s_i \in X^s$ for simplicity. 

We assume that, within each source $s$, all labeled examples are generated according to the same distribution, and likewise for the target. We further assume that each domain is generated by a different distribution, but that they are similar enough that a model trained on one can be transferred, to some degree, to another. Namely, as discussed in Section~\ref{section:intro}, we suppose that the target domain can be made to resemble any fixed source domain $(X^s, Y^s)$ via a transformation $\phi_s$. Once transformed, samples from the target domain could be classified using a model trained on the $s$-th source domain. The key question then is how to find such a transformation.

As a natural starting point, we first train a classifier $f_s$ to label samples from source dataset $(X^s, Y^s)$. Because the distributions generating the source and target are different, the accuracy of $f_s$ on the target $X^\star$ will tend to be much worse than that of $f_s(X^s)$. 
The hope is that predictions obtained by transforming the target samples before classification---that is, $f_s(\phi_s(X^\star))$---will yield more favorable results.

In addition to the classifier $f_s$, we suppose the availability of an approximate labeling $\hat{Y}^\star = \langle \hat{y}^\star_1, \ldots, \hat{y}^\star_n \rangle$, $\hat{y}^\star_i \in [0, 1]$ on the target samples $X^\star$. Such a rough labeling could come from a variety of simple learning procedures, and we outline some options in Section~\ref{sec:roughLabeling}. To estimate the transformation $\phi_s$, we leverage both $f_s$ and the approximate labeling  $\hat{Y}^\star$. 

Specifically, if $f_s$ is the ideal decision boundary for an optimally-transformed target, and $\hat{Y}^\star$ is a labeling on the target that is usually correct, then we may be able to uncover the optimal transformation as the function $\phi_s$ that causes $\hat{Y}^\star$ to best agree with $f_s$'s predictions on the transformed target samples $f_s(\phi_s(x^\star_i))$. That is, we find $\phi_s$ to minimize the squared error: 
\begin{equation}
\label{eq1}
\phi_s = \arg\min_{\phi} \sum_{i=1}^{n} \big(\hat{y}^\star_i - f_s(\phi(x^\star_i))\big)^2.
\end{equation}
The nested nature of $\phi$ makes gradient-based methods like neural networks an ideal model choice for this optimization.

We refer to this procedure, which treats each source separately, as \emph{approximate label matching}. When there are multiple sources available ($k > 1$), we can run ALM on each to create multiple predictors, then take a simple average:
\begin{equation}
\label{eq2}
H(x^\star_i) = {\rm round} \Big(\frac{1}{k} \sum_{s=1}^{k} f_s(\phi_s(x^\star_i))\Big).
\end{equation}
This average hypothesis can be similarly obtained for any domain adaptation algorithm.

Our method can be generalized to multiclass problems by using a vector encoding for labels and predictions instead of scalar values.

\subsection{Obtaining a Rough Labeling}
\label{sec:roughLabeling}

As discussed above, approximate label matching requires a better-than-guessing estimate $\hat{Y}^\star$ of the target domain labels in order to be effective. In this section, we overview two ways in which these approximate labels can be acquired.

In the first approach, we obtain a rough labeling by treating all sources in aggregate as one larger training set, and fitting a classifier as one would in an ordinary supervised (non-domain-adaptation) setting. We could then use the predictions of this classifier on the target samples as $\hat{Y}^\star$. We call this a \emph{pseudo-supervised} rough labeling. This works particularly well when there are multiple sources available, because the resulting model will be able to leverage features that generalize well across the different sources, and that may also generalize reasonably well to an unobserved target domain.

Another approach for estimating target domain labels, which may be preferred when only one source dataset is available, is to simply obtain a hypothesis from an alternative unsupervised domain adaptation algorithm, and use it to produce a rough labeling on target samples. When using ALM in conjunction with this kind of rough labeling, we call the overall technique \emph{refinement}, since ALM is in a sense refining the labeling produced by the given domain adaptation algorithm.

\subsection{Adversarial Regularization}
\label{sec:adreg}
Approximate label matching attempts to align the target $X^\star$ with the source $X^s$. However, when the transformation $\phi_s$ is highly expressive, we may find that it contorts the target samples so as to match the rough labeling $\hat{Y}^\star$ without aligning $X^\star$ to $X^s$. We have observed this phenomenon anecdotally and frequently in experiments; a detailed example is given in Section~\ref{section:synth}. This is a form of overfitting, in the sense that ALM is finding a hypothesis that agrees with $\hat{Y}^\star$ too precisely due to underconstrained optimization.

In order to prevent this type of overfitting, we propose an adversarial form of regularization, which attempts to force the transformed target samples $\phi_s(X^\star)$ to look enough like the source samples $X^s$ that they could confuse a classifier trained to distinguish between the two. We do this by adding a discriminating function $D_s$ that is trained jointly with the other ALM machinery. At each step of learning, $D_s$ is updated to better distinguish between source samples and transformed target samples, and $\phi_s$ is updated to both deteriorate the performance of $D_s$ and to solve Eqn.~(\ref{eq1}). Our regularization technique is very similar to traditional Generative Adversarial Networks (GANs) \cite{gan}, but instead of generating samples from random noise, we generate source-looking samples from target samples. The discriminating function takes as input either source samples or transformed target samples, and is trained to predict either 0 for a source sample or 1 for a target sample. The loss minimized by $D_s$ is simply the binary cross-entropy of the misclassified source and target samples,
\begin{eqnarray*}
-\sum_{i=1}^{n} \lambda  \log \big(D_s(\phi_s(x^\star_i)) \big) - \sum_{i=1}^{m_s} \lambda \log \big(1 - D_s(x^s_i) \big),
\end{eqnarray*}

where $\lambda \in [0, 1]$ is the magnitude of the regularization. This approach changes the loss function for $\phi_s$ to 
\begin{eqnarray*}
\sum_{i=1}^{n} \big[ (1 - \lambda)  \big(\hat{y}^\star_i - f_s(\phi_s(x^\star_i))\big)^2 +   \lambda \mathcal{L}(D_s(\phi_s(x^\star_i)), 1)) \big],
\end{eqnarray*}
 where $\mathcal{L}(a,b)$ is the binary cross-entropy loss resulting from output $a$ with label $b$. The second term,  $\mathcal{L}(D_s(\phi_s(x^\star_i)), 1))$, denotes the discriminating function's confidence that the transformed target sample is a target sample. Minimizing this confidence corresponds to ``confusing'' the discriminator, effectively making the transformed target samples resemble source samples. The fitted ALM now needs to match the labeling $\hat{Y}^\star$ as well as possible, while also aligning $X^\star$ to source $X^s$.

Adversarial regularization makes the source and transformed target samples appear relatively similar without the need to approximate generative distributions or define an explicit distance metric between them. 

\subsection{Implementation of ALM}

We implement the ALM functions using neural networks. Specifically, we let $f_s$ be a neural network of arbitrary architecture, and we then use this network to estimate $\phi_s$. When solving for $\phi_s$, $f_s$ and $\hat{Y}^\star$ have already been approximated and are fixed. The transformation $\phi_s$ is represented by adding additional network layers before the existing input layer of the already-trained $f_s$ network. We further add a discriminator $D_s$ to the output of $\phi_s$, which, as described earlier, is iteratively trained to distinguish between source samples and transformed target samples.

\begin{figure}[h]
\vskip 0.2in
\begin{center}
\centerline{\includegraphics[trim={4cm 8cm 12cm 10cm}, clip,scale=0.33]{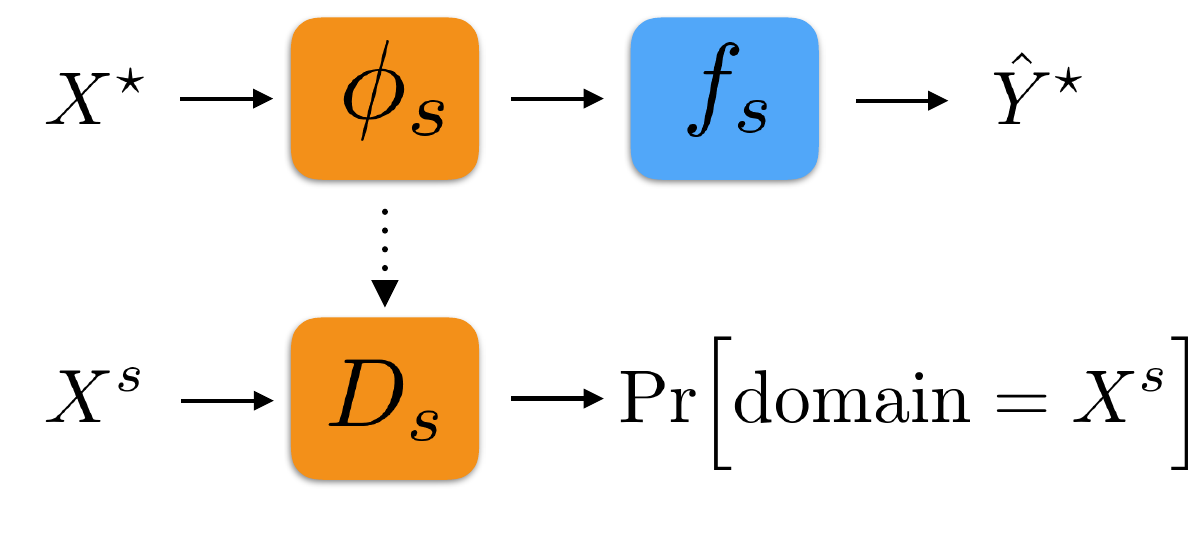}}
\caption{\textbf{A diagram of ALM for a fixed source $s$.} Orange modules are being trained together while blue modules were trained offline and are fixed. Solid arrows correspond to standard connections and dashed arrows are adversarial connections.}
\label{architecture}
\end{center}
\vskip -0.2in
\end{figure} 

When training this aggregate network, we use $\hat{Y}^\star$ as labels on $X^\star$ for backpropagation with a mean squared error criterion, and we do not update any of the weights in the layers corresponding to $f_s$. The compositionality of backpropagation allows us to easily solve Eqn.~(\ref{eq2}) even though $\phi_s$ is nested within $f_s$.

\section{A Synthetic Example}
\label{section:synth}
\begin{figure}[h]
\vskip 0.2in
\begin{center}
\centerline{\includegraphics[trim={0cm 0cm 2cm 0cm}, scale=0.28]{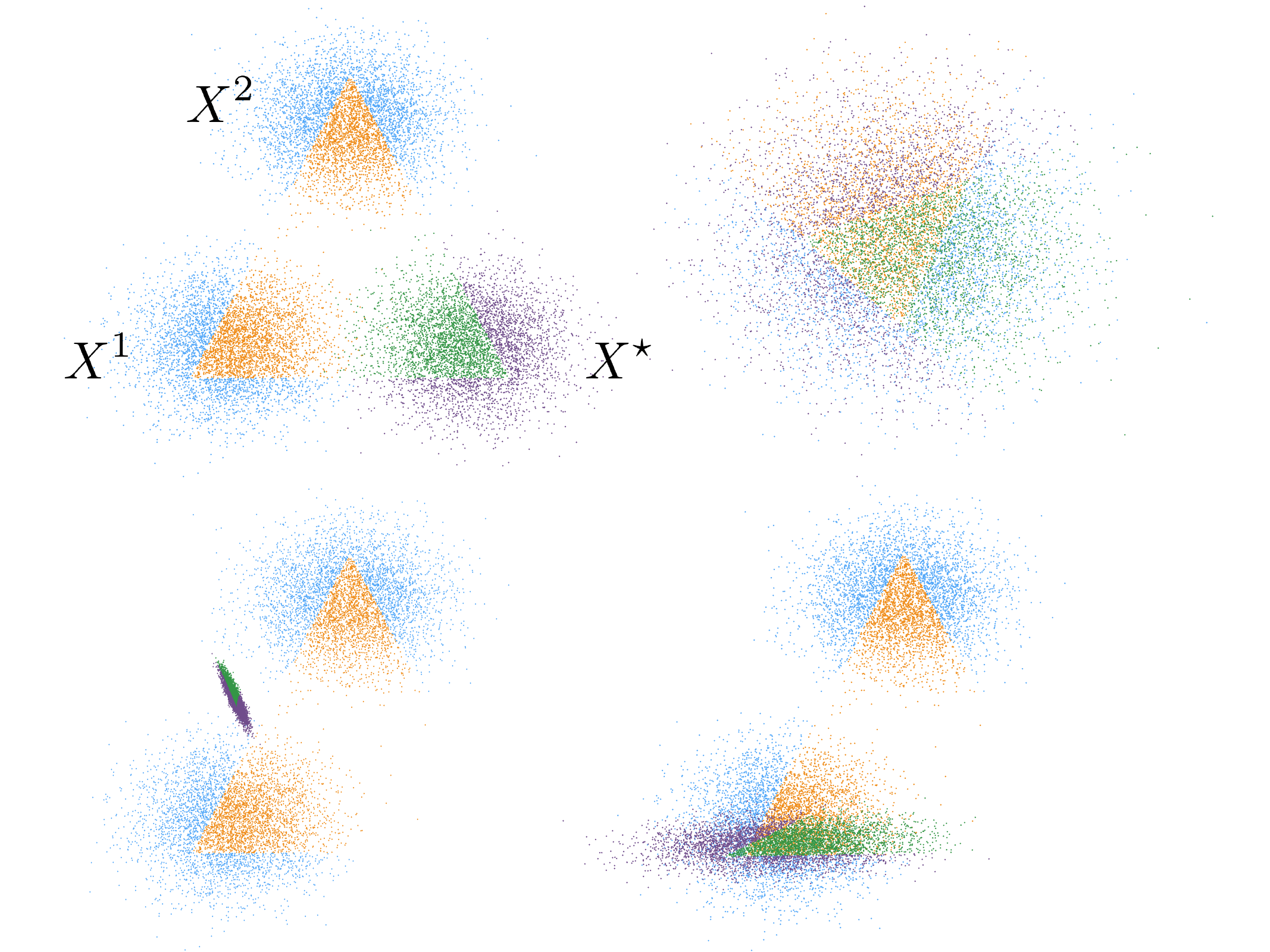}}
\caption{\textbf{A visualization of the synthetic dataset and a few domain adaptation techniques.} \textbf{Top Left:} The original dataset as shown in Figure~\ref{ex0}. \textbf{Top Right:} The source $X^1$ and target $X^\star$ after being transformed by SA. \textbf{Bottom Left:} The transformation learned by ALM without adversarial regularization applied to the target data. \textbf{Bottom Right:} The transformation learned by ALM with adversarial regularization applied to the target data.}
\label{synthEx}
\end{center}
\vskip -0.2in
\end{figure}

To illustrate how our algorithm works, we demonstrate its performance on the artificial dataset described in Section~\ref{section:intro}, which is difficult to classify but conceptually simple.
The data consists of three isotropic Gaussian domains distributed on each of the three vertices of an equilateral triangle (Figure~\ref{synthEx}). Points that are inside the triangle's perimeter are assigned negative labels, and those outside are assigned positive labels. For this example, we will try to adapt the target domain $X^\star$ to the first source domain $X^1$, but performance is similar for any source-target pair chosen.

Because our example includes two sources, $\hat{Y}^\star$ is obtained by training a multi-layer perceptron (MLP) on both sources simultaneously. Such a classifier is able to correctly predict 79.9\% of the samples in the held-out target domain.

ALM trained without adversarial regularization overfits in exactly the way described in Section~\ref{sec:adreg}: It learns to make predictions that almost exactly matches the rough labeling $\hat{Y}^\star$, but contorts $X^\star$ such that it does not resemble $X^1$ at all (Figure~\ref{synthEx}).
Once adversarial regularization is included, the learned transformation $\phi_1$ aligns $X^\star$ to $X^1$ fairly well, yielding 90.0\% accuracy.  

This unsupervised domain adaptation simulation demonstrates ALM's ability to find transformations that other algorithms cannot. In this specific example, recovering the correct translation of target to source inputs is straightforward even without using a target labeling, but recovering the correct domain rotation is challenging because of the circular shape of the datasets. An algorithm that does not make use of a target prediction cannot be expected to perform well on this task.

As an example, the SA algorithm, which does not use a target labeling, lumps the source and target together so that they appear to be from the same distribution, but cannot uncover the necessary target domain rotation (Figure~\ref{synthEx}). Accordingly, SA does not perform better than guessing on this adversarial simulation.

\section{Experiments and Results}
\label{section:results}

We turn next to an empirical evaluation of ALM compared to some leading domain adaptation algorithms. We evaluate these methods on several domain adaptation problems, which are described in detail below. We begin by discussing our experimental setup.

\subsection{Experimental Setting}

Each domain-adaptation problem consists of several domains. In our experiments we use one domain as a target and another as a source. We repeat this for all possible source-target pairs and report accuracies for each. 

In each experiment, we chose $\phi_s$ to be an MLP that includes three layers, allowing for nonlinear transformations, with 50 hidden nodes and input and output sizes equal to the dimensionality of the data. Similarly, we chose $f_s$ and the predictor used to produce pseudo-supervised rough labelings to both be MLPs with three layers, a hidden layer size of 50 nodes, and as many output dimensions as there are classes in the problem. We fix $D_s$ to also be an MLP with three layers, and again with 50 nodes in the hidden layer. In all experiments, dropout 50\% is applied to help in training all networks except for $\phi_s$. 

We compare ALM to both DANNs and SA on all datasets. The SA algorithm takes a source and target as inputs and returns a canonical feature representation. To make a reasonable comparison, once in this space, we use a neural network with the same architecture as used for $f_s$ in ALM for classification. Similarly, for DANNs, we fix the transformation, label classification, and domain classification modules to have the same architecture as $\phi_s$, $f_s$, and $D_s$, respectively. While these architectures are certainly not designed to be optimal for each application, we kept them fixed as a control to highlight the relative performance of the various domain adaptation approaches and the generality of our solution.

We also include accuracies of the approximate labeling $\hat{Y}^\star$ as well as $f_s(X^\star)$, that is, $f_s$ applied to the target directly, without transformation by $\phi_s$. Finally, we report the results of a ``target only'' experiment (denoted TO in the tables), in which the target domain is used for both training and testing (still using the same architecture as $f_s$). These accuracies are computed using twenty-fold cross validation. The purpose of this design is to show for comparison how well a model can perform when training and test examples come from the same distribution.

The domain adaptation techniques used here, especially DANNs and ALM, can be hyperparameter sensitive. The unsupervised approximation of these kinds of values is an active area of research outside the scope of this work. To obtain accuracy values, we ran each algorithm 100 times with randomly-chosen hyperparameters for each source-target pair, and reported the best results. As an aside, we did notice that the value of $D_s(\phi_s(X^\star))$, the discriminator's confidence that the transformed target samples are indeed target samples, tends to correlate strongly with accuracy in both ALM and DANN: exploiting this observation for hyperparameter tuning would be an interesting direction for future research. All neural networks were implemented in Torch \cite{torch} and optimized using the Adam variant of SGD \cite{adam}.

\subsection{Sentiment Classification}

In the sentiment dataset, we are provided with product reviews from {\tt amazon.com} \cite{sentiment}. All reviews come from items in either Amazon's book, DVD, electronics, or kitchen departments, and data from each department contains 2,000 samples; each department is a separate domain. Each review is represented as term frequencies for a specified vocabulary. The goal is to distinguish between positive and negative reviews.

Because there are multiple sources available in this problem, we used a pseudo-supervised rough labeling (see Section~\ref{sec:roughLabeling}) for ALM. However, because of this choice, for a fixed source-target pair, $\hat{Y}^\star$ implicitly uses more information than DANNs and SA, which only use information from that single source. Thus, we also present results with refinement of SA and DANN labelings.

We find that, for this task, DANN generally outperforms SA, sometimes by a large margin (Table~\ref{err-senti}). ALM consistently achieves a higher accuracy than the rough labeling used during its optimization, regardless of the procedure used to generate that labeling.

\newcolumntype{C}[1]{>{\centering\arraybackslash}p{#1}}
\newcolumntype{L}[1]{>{\raggedleft\arraybackslash}p{#1}}
\newcolumntype{R}[1]{>{\raggedright\arraybackslash}p{#1}}
	
\begin{table}[h]
\caption{{\bf Percent accuracies for the sentiment classification task.} Here, D, B, E, and K represent the DVD, Book, Electronics, and Kitchen Appliances domains respectively. Pairs of letters (e.g., DB) correspond to using the \emph{DVD} domain as a source for labeling the \emph{Books} domain, which is used as a target. ALM\textsubscript{$p$}, ALM\textsubscript{$s$}, and ALM\textsubscript{$d$} show the result of performing ALM using a pseudo-supervised rough labeling $\hat{Y}^\star$, refinement of SA, and refinement of DANN, respectively. Average accuracies are shown in the last row.}
\label{err-senti}
\vskip 0.15in
\begin{center}
\begin{small}
\begin{sc}
\resizebox{\columnwidth}{!}{
\begin{tabular}{lcccccccr}
\hline
\abovespace\belowspace
 &  TO &  $f_s(X^\star)$ & $\hat{Y}^\star$ & ALM\textsubscript{$p$} &  DANN &  ALM\textsubscript{$d$} &  SA & ALM\textsubscript{$s$}\\
\hline
\abovespace
DB & 85.5 & 54.6 & 80.3 & \textbf{82.9} & 76.8 & 78.4 & 74.8 & 77.0\\
EB & 85.5 & 52.0 & 80.3 & \textbf{82.8} & 63.1 & 67.1 & 65.2 & 66.0\\
KB & 85.5 & 55.6 & 80.3 & \textbf{82.7} & 65.1 & 65.7 & 67.6 & 70.5\\
BD & 82.3 & 55.0 & 80.2 & \textbf{82.7} & 80.0 & 81.2 & 74.5 & 76.4\\
ED & 82.3 & 53.9 & 80.2 & \textbf{83.2} & 66.1 & 68.0 & 63.9 & 64.3\\
KD & 82.3 & 55.6 & 80.2 & \textbf{82.6} & 67.2 & 68.4 & 60.1 & 61.4\\
BE & 81.8 & 50.7 & 82.8 & \textbf{85.2} & 68.9 & 69.1 & 62.0 & 66.6\\
DE & 81.8 & 54.4 & 82.8 & \textbf{84.9} & 70.2 & 70.3 & 64.8 & 65.0\\
KE & 81.8 & 65.8 & 82.8 & \textbf{85.1} & 76.2 & 79.1 & 54.3 & 56.2\\
BK & 80.0 & 53.6 & 87.0 & \textbf{89.0} & 75.1 & 75.4 & 64.4 & 68.6\\
DK & 80.0 & 54.4 & 87.0 & \textbf{89.2} & 71.5 & 72.2 & 56.6 & 60.0\\
EK & 80.0 & 60.9 & 87.0 & \textbf{89.2} & 76.2 & 78.7 & 55.6 & 56.7\\
\hline
\belowspace
    & 82.4 &  55.4 & 82.6 & \textbf{85.0} & 76.2 &  72.8 & 63.7 & 65.7\\
\hline
\end{tabular}}
\end{sc}
\end{small}
\end{center}
\vskip -0.1in
\end{table}

\subsection{Digit Classification}

\begin{figure}[h]
\vskip 0.2in
\begin{center}
\centerline{\includegraphics[trim={5cm 12cm 5cm 13cm}, clip, scale=0.7]{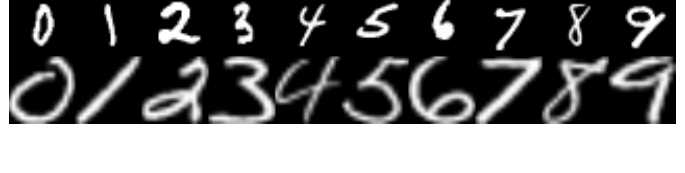}}
\caption{{\bf Example MNIST and USPS digit images.} The top row shows MNIST images and the bottom row shows USPS images. The USPS images are actually 16x16 pixels, so they were scaled to the same size as MNIST for this figure.}
\label{mnistusps}
\end{center}
\vskip -0.2in
\end{figure}

In this experiment, we are supplied with standard MNIST digits, as well as handwritten binary digits extracted from USPS parcels. These two datasets look very similar (Figure~\ref{mnistusps}), but a classifier trained on one still performs significantly worse on the other. Digits from the USPS domain are 16 pixels smaller in both dimensions than MNIST images, so we zero-pad them by 8 pixels on each side for preprocessing.

Because only one source is available, we use refinement of SA to acquire rough labels $\hat{Y}^\star$. To obtain image features, we convolutionally embed both datasets into 84 dimensions. This is done by training a convolutional network of the LeNet-5 \cite{lenet} architecture on the source data, then taking the activations in the layer before the output layer as a new representation of both the source and the target (Table~\ref{err-digit}). \looseness=-1

\begin{table}[h]
\caption{{\bf Percent accuracies for the digit classification task.} The first row of results correspond to treating USPS images as a target and the second row corresponds to using MNIST images as a target. Average accuracies are shown in the last row.}
\label{err-digit}
\vskip 0.15in
\begin{center}
\begin{small}
\begin{sc}
\begin{tabular}{lccccr}
\hline
\abovespace\belowspace
 & TO & $f_s(X^\star)$ & ALM & DANN & SA\\
\hline
\abovespace
U &90.8 & 90.0& \textbf{94.5}& 92.4 & 93.1\\
M &88.0 & 86.9& \textbf{89.7}& 87.8 & 88.3\\
\hline
\belowspace
   & 89.2 & 88.5& \textbf{92.2}& 90.1 & 90.7\\
\hline
\end{tabular}
\end{sc}
\end{small}
\end{center}
\vskip -0.1in
\end{table}

Our results show that even in a situation where the source and target appear very similar, domain adaptation could be used to improve test accuracy. We find that ALM achieves a more favorable result than that produced by other techniques on these digit datasets. \looseness=-1 

\subsection{Office Object Classification}

\begin{figure}[h]
\vskip 0.2in
\begin{center}
\centerline{\includegraphics[trim={0cm 4cm 0cm 0cm}, clip, scale=0.28]{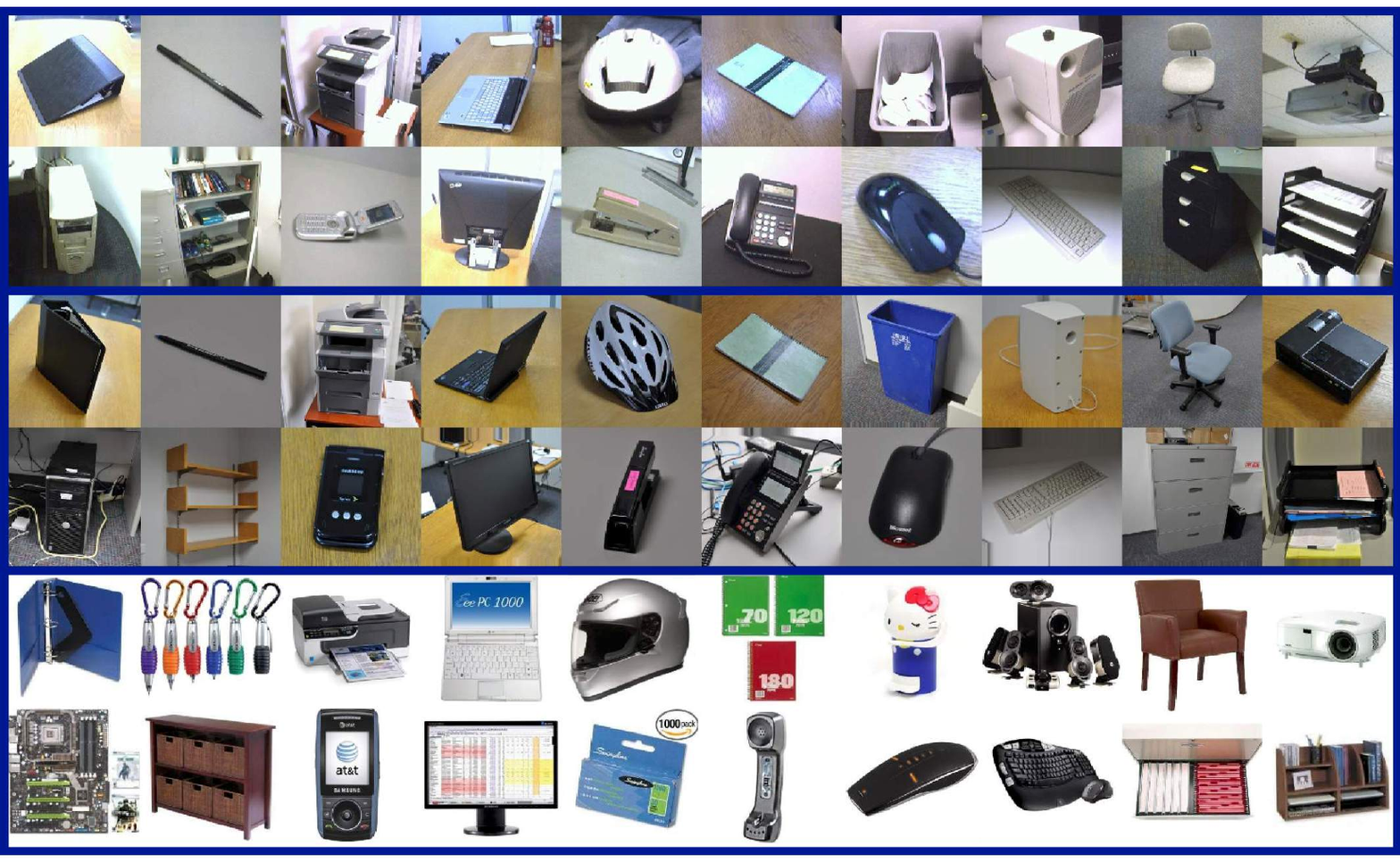}}
\caption{{\bf Example images for 20 (randomly chosen) of the 31 classes in the Office Dataset.} Classes from top left to bottom right are \emph{ring binder, pen, printer, laptop computer, helmet, paper notebook, trash can, speaker, desk chair, projector, desktop computer, bookshelf, mobile phone, monitor, stapler, phone, mouse, keyboard, file cabinet}, and \emph{letter tray}. The top two rows of images are from the \emph{webcam} domain, and tend to look overexposed and washed-out, while more balanced images on the third and fourth rows are generated from the \emph{DSLR} domain. The last two rows of images correspond to Amazon product images, which generally feature a white background.}
\label{stanford}
\end{center}
\vskip -0.2in
\end{figure}

The Stanford Office dataset consists of images of 31 different objects typically found around an office \cite{officedata}. It is composed of three domains, pictures taken by a low-quality webcam, a high-quality DSLR camera, and product images from {\tt amazon.com} (Figure~\ref{stanford}).

The webcam, DSLR, and Amazon domains consist of 795, 498, and 2,817 images, respectively. Because state-of-the-art image classifiers typically require much more data to train than is available here, we use a network of the AlexNet \cite{alexnet} architecture that has been trained on the ImageNet dataset as a feature extractor. As in the previous experiment, we pass each image through the network and use the activations at the last hidden layer as a lower-dimensional representation of the data. In this section, we use refinement of SA for the rough labeling $\hat{Y}^\star$. \looseness=-1

\begin{table}[h]
\caption{\textbf{Percent accuracies for the object classification task.} Here, A, D, and W represent the \emph{Amazon}, \emph{DSLR}, and \emph{webcam} domains respectively. Paired notation (e.g., DA) corresponds to using the \emph{DSLR} domain as a source for labeling target \emph{Amazon} domain. Average accuracies are shown in the last row.}
\label{err-stanford}
\vskip 0.15in
\begin{center}
\begin{small}
\begin{sc}
\begin{tabular}{lccccr}
\hline
\abovespace\belowspace
 & TO & $f_s(X^\star)$ & ALM & DANN & SA\\
\hline
\abovespace
DA & 69.5 & 30.1 & \textbf{35.9} & 35.3 & 35.2\\
WA & 69.5 & 32.2 & \textbf{37.5} & \textbf{37.5} & 36.9\\
AD & 77.4 & 42.0 & \textbf{53.6} & 47.0 & 49.4\\
WD & 77.4 & 94.1 & \textbf{98.0} & 94.4 & 96.4\\
AW & 80.0 & 38.3 & \textbf{48.1} & 43.8 & 45.9\\
DW & 80.0 & 80.4 & \textbf{90.8} & 86.0 & 88.3\\
\hline
\belowspace
&75.6 & 52.9 & \textbf{60.7} & 57.3 &	58.7\\
\hline
\end{tabular}
\end{sc}
\end{small}
\end{center}
\vskip -0.1in
\end{table}

In these results, the most challenging source-target pairs are those that involve the Amazon domain (Table~\ref{err-stanford}). Unlike the webcam and DSLR domains, Amazon images could contain more than a single object, and that object may not resemble typical items found around an office. Even in these difficult situations, we find that ALM performs as well or better than other methods. 

We note that these results are somewhat worse than those reported in the original DANN article. In that work, a pretrained AlexNet is refined during training as the transformation module, rather than used to produce an image embedding and training a transformation module from scratch as we do here \cite{dannJmlr}. Although not explored in this work, we would expect a similar boost in performance from using the alterantive approach with ALM.

\subsection{MEG Signal Classification}
\label{section:meg}

Magnetoencephalography (MEG) machines use magnetic fields to measure the brain's electrical activity and are frequently used for experiments in neuroscience and psychology. In the specific problem we consider, originally part of an online competition, the data come from sixteen different people who were shown either pictures of human faces or non-faces (images with no discernible face-like structures) while having their brain activities recorded using MEG. Each subject's data comprises a single domain, and the objective is to classify instances where a held-out subject was looking at a face from those in which the subject was looking at a non-face \cite{meg}.
  
Domain adaptation is a problem that comes up frequently in brain-computer interface problems in general, not just those dealing with MEG data. Advancing classification methods in this space could improve existing techniques in areas like fMRI analysis, neuropsychiatric disease diagnosis, prosthetics, and human-computer interaction \cite{bciev}, where the locations of neurons and voxels are often person-specific \cite{voxel}.
 
\begin{figure}[h]
\vskip 0.2in
\begin{center}
\centerline{\includegraphics[trim={0cm 0cm 0cm 0cm}, clip, scale=0.23]{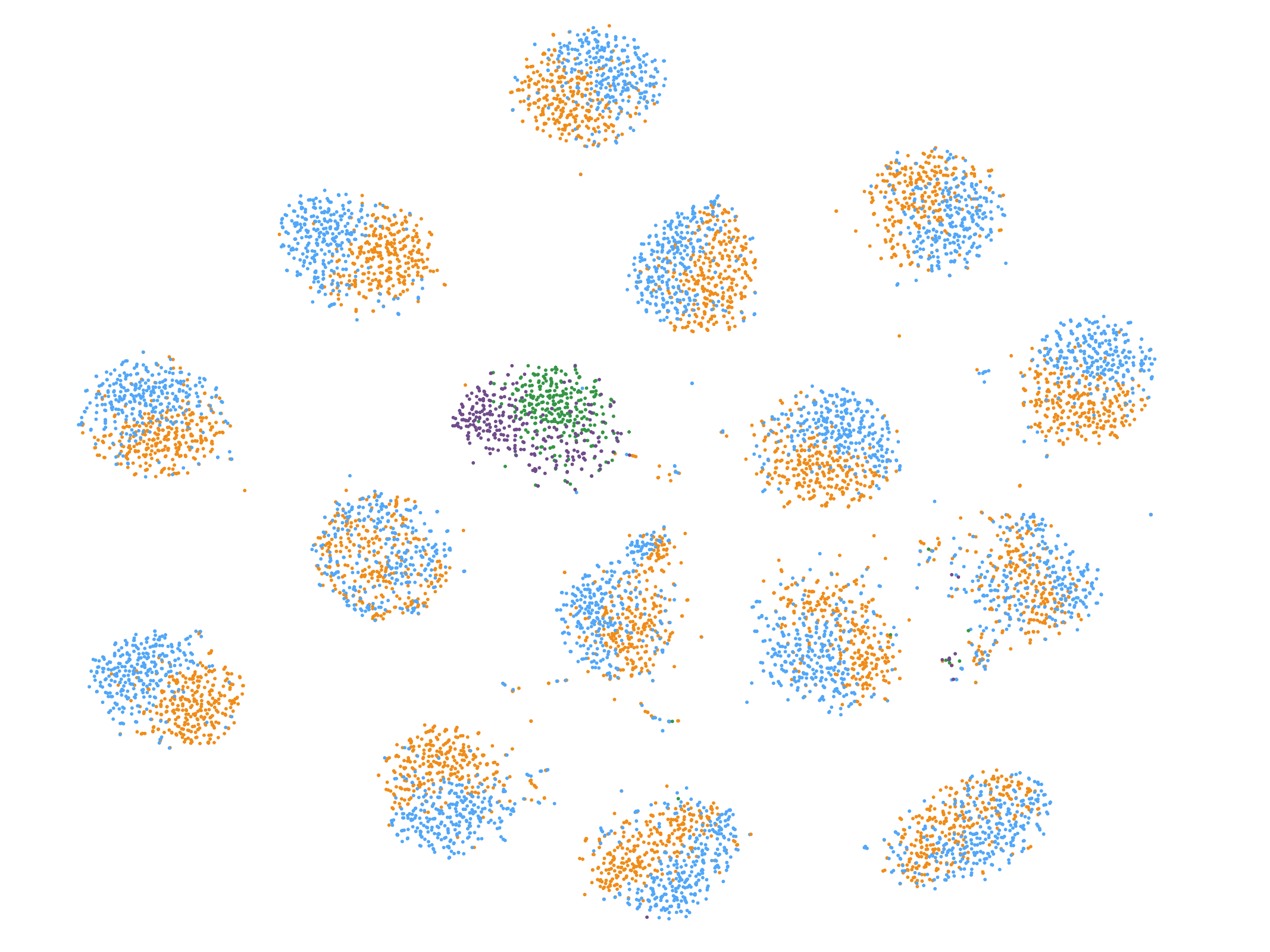}}
\caption{\textbf{A visualization of the MEG data embedded into two dimensions using t-SNE \cite{tsne}.} Blue points are positive source samples, orange points are negative source samples, and purple and green points are positive and negative target samples respectively. The data shows distinct clusters corresponding to each subject, making this a clear domain adaptation problem. Target domain labels are shown here for clarity, but the classifier does not have access to labels from the target dataset.}
\label{megFig}
\end{center}
\vskip -0.2in
\end{figure}

We represent this data according to the method discussed by \citet{megRep}, which maps the raw time-series data into 2,014 dimensions. Source labels are used to obtain this representation, so each held-out subject operates in a different space. Figure~\ref{megFig} shows an example representation embedded into two dimensions. Each domain contains between 500 and 600 samples. \looseness = -1

These data, at least in an embedded space, seems to have visual properties similar to the synthetic example discussed earlier. Most domains appear to be rotated and translated with respect to each other, and there is no geometric overlap between them. This suggests that ALM might be particularly well-suited for this problem.
 
Unlike in previous experiments, instead of reporting the accuracy of a technique that performs domain adaptation on a fixed source and target, we average the hypotheses resulting from running that technique on each available source and a fixed target. For ALM, this average hypothesis is described by Eqn.~(\ref{eq2}). In this problem we again use a pseudo-supervised rough labeling as $\hat{Y}^\star$.

We also compare to stacked generalization, which in the MEG literature is often used to solve such problems \cite{stack}. In this hierarchical approach, a separate base classifier is trained on each of the $k$ available sources. The predictions of these $k$ classifiers on all available source samples are then aggregated and used as a $k$-dimensional feature space for a higher-level classifier. Again, all models are MLPs with one hidden layer composed of 50 nodes.\looseness=-1

\begin{table}[t]
\caption{\textbf{Percent accuracies for the MEG classification task.} Each row presents results corresponding to holding out the denoted subject as a target and averaging the hypotheses resulting from performing domain adaptation with each available source. For this experiment we also include the results of stacked generalization, shown below as SG. Average accuracies are shown in the last row.}
\label{err-meg}
\vskip 0.15in
\begin{center}
\begin{small}
\begin{sc}
\resizebox{\columnwidth}{!}{
\begin{tabular}{lccccccr}
\hline
\abovespace\belowspace
 & TO & $f_s(X^\star)$ & $\hat{Y}^\star$ & ALM & DANN & SA & SG\\
\hline
\abovespace
S01 & 85.6 & 75.1 & 77.2 & \textbf{85.2}& 77.1 & 76.3 & 72.7\\
S02 & 80.7 & 66.4 & 64.8 & \textbf{74.7}& 69.9 & 66.8 & 62.1\\
S03 & 82.9 & 59.8 & 61.1 & \textbf{73.0}& 65.2 & 64.0 & 57.1\\
S04 & 90.5 & 72.2 & 72.2 & \textbf{90.6}& 79.9 & 75.9 & 80.4\\
S05 & 86.2 & 66.5 & 66.0 & \textbf{82.5}& 76.1 & 70.8 & 62.9\\
S06 & 86.1 & 69.0 & 66.1 & \textbf{75.7}& 70.4 & 73.1 & 63.4\\
S07 & 86.1 & 53.9 & 69.9 & \textbf{83.0}& 68.7 & 70.0 & 53.7\\
S08 & 87.3 & 70.8 & 66.8 & \textbf{82.4}& 76.8 & 71.9 & 64.9\\
S09 & 88.2 & 73.7 & 71.7 & \textbf{84.1}& 79.4 & 73.7 & 76.1\\
S10 & 87.7 & 64.0 & 68.6 & \textbf{81.4}& 75.0 & 71.3 & 60.5\\
S11 & 78.7 & 62.5 & 67.4 & \textbf{72.5}& 64.4 & \textbf{72.5} & 59.3\\
S12 & 84.0 & 78.4 & 76.9 & \textbf{86.2}& 76.4 & 74.7 & 75.1\\
S13 & 83.3 & 69.9 & 67.8 & \textbf{79.3}& 75.0 & 72.9 & 70.2\\
S14 & 91.4 & 73.6 & 74.5 & \textbf{88.6}& 79.6 & 74.4 & 72.9\\
S15 & 90.7 & 65.0 & 68.2 & \textbf{83.6}& 72.7 & 70.7 & 67.4\\
S16 & 88.2 & 56.1 & 63.0 & \textbf{78.0}& 66.1 & 67.4 & 58.6\\
\hline
\belowspace
   & 86.1 & 67.3 &	68.9 & \textbf{81.3} & 73.3 &  71.6 &	66.1\\
\hline
\end{tabular}}
\end{sc}
\end{small}
\end{center}
\vskip -0.1in
\end{table}

ALM outperforms other techniques on this data by a margin larger than in any of the preceding experiments (Table~\ref{err-meg}).

\section{Discussion and Future Work}

In this article, we presented approximate label matching, a novel and powerful technique for unsupervised domain adaptation. 

In our results, we found that the performance of ALM is never worse than that of the rough labeling used in its optimization. In several cases, ALM was only able to improve $\hat{Y}^\star$ slightly, highlighting the difficulty of the unsupervised domain adaptation task. Still, our approach can be used in conjunction with any domain adaptation technique. In our results, ALM outperformed other state-of-the-art domain adaptation methods, sometimes by a substantial margin. \looseness=-1

Separately, it might be useful to apply this algorithm iteratively to see if performance continues to improve.
In that setting, the output of the current iteration of ALM would be the approximate labeling used for the next iteration of ALM. \looseness=-1

Although we only address unsupervised domain adaptation in this article, our approach could be straightforwardly modified to accommodate supervised or semi-supervised situations as well. For supervised domain adaptation, we could learn a transformation by simply setting $\hat{Y}^\star$ to the true labels of the target. Similarly, in semi-supervised situations, the available target labels could be used to supplement the source data when learning $\hat{Y}^\star$.\looseness=-1

\bibliography{main}
\bibliographystyle{icml2017}
\end{document}